\documentclass[a4paper,11pt]{article}
\usepackage[margin=1in]{geometry}
\usepackage{times}  
\usepackage{helvet}  
\usepackage{courier}  
\usepackage[hyphens]{url}  
\usepackage{graphicx} 
\usepackage{natbib}  
\usepackage{caption} 
\usepackage{algorithm}
\usepackage{algorithmic}

\usepackage{amsmath} 
\usepackage{booktabs} 
\usepackage[many]{tcolorbox}
\usepackage{multirow}
\usepackage{cleveref}
\usepackage{lineno}

\usepackage{newfloat}
\usepackage{listings}

\title{HMR3D:Hierarchical Multimodal Representation for 3D Scene Understanding with Large Vision-Language Model}

\author{
Chen Li\textsuperscript{1,2}\thanks{Equal contribution.}\,
\and
Eric Peh\textsuperscript{1,2}\footnotemark[1]\,
\and
Basura Fernando\textsuperscript{1,2,3}\\
\textsuperscript{1} Institute of High-Performance Computing, Agency for Science, Technology and Research, \\ Singapore \\
\textsuperscript{2} Centre for Frontier AI Research, Agency for Science, Technology and Research, Singapore \\
\textsuperscript{3} College of Computing and Data Science, Nanyang Technological University, Singapore
}

\usepackage{bibentry}

\begin{document}

\date{}

\maketitle

\begin{abstract}
Recent advances in large vision-language models (VLMs) have shown significant promise for 3D scene understanding. Existing VLM-based approaches typically align 3D scene features with the VLM’s embedding space. However, this implicit alignment often yields suboptimal performance due to the scarcity of 3D data and the inherent complexity of spatial relationships in 3D environments. To address these limitations, we propose a novel hierarchical multimodal representation for 3D scene reasoning that explicitly aligns with VLMs at the input space by leveraging both multi-view images and text descriptions. The text descriptions capture spatial relationships by referencing the 3D coordinates of detected objects, while the multi-view images include a top-down perspective and four directional views (forward, left, right, and backward), ensuring comprehensive scene coverage. Additionally, we introduce a hierarchical feature representation that aggregates patch-level image features into view-level and scene-level representations, enabling the model to reason over both local and global scene context. Experimental results on both situated 3D Q\&A and general 3D Q\&A benchmarks demonstrate the effectiveness of our approach. Code will be released publicly.
\end{abstract}

\section{Introduction}
\label{sec:intro}

\begin{figure}[tbp]
\centering
    \includegraphics[width=0.45\textwidth]{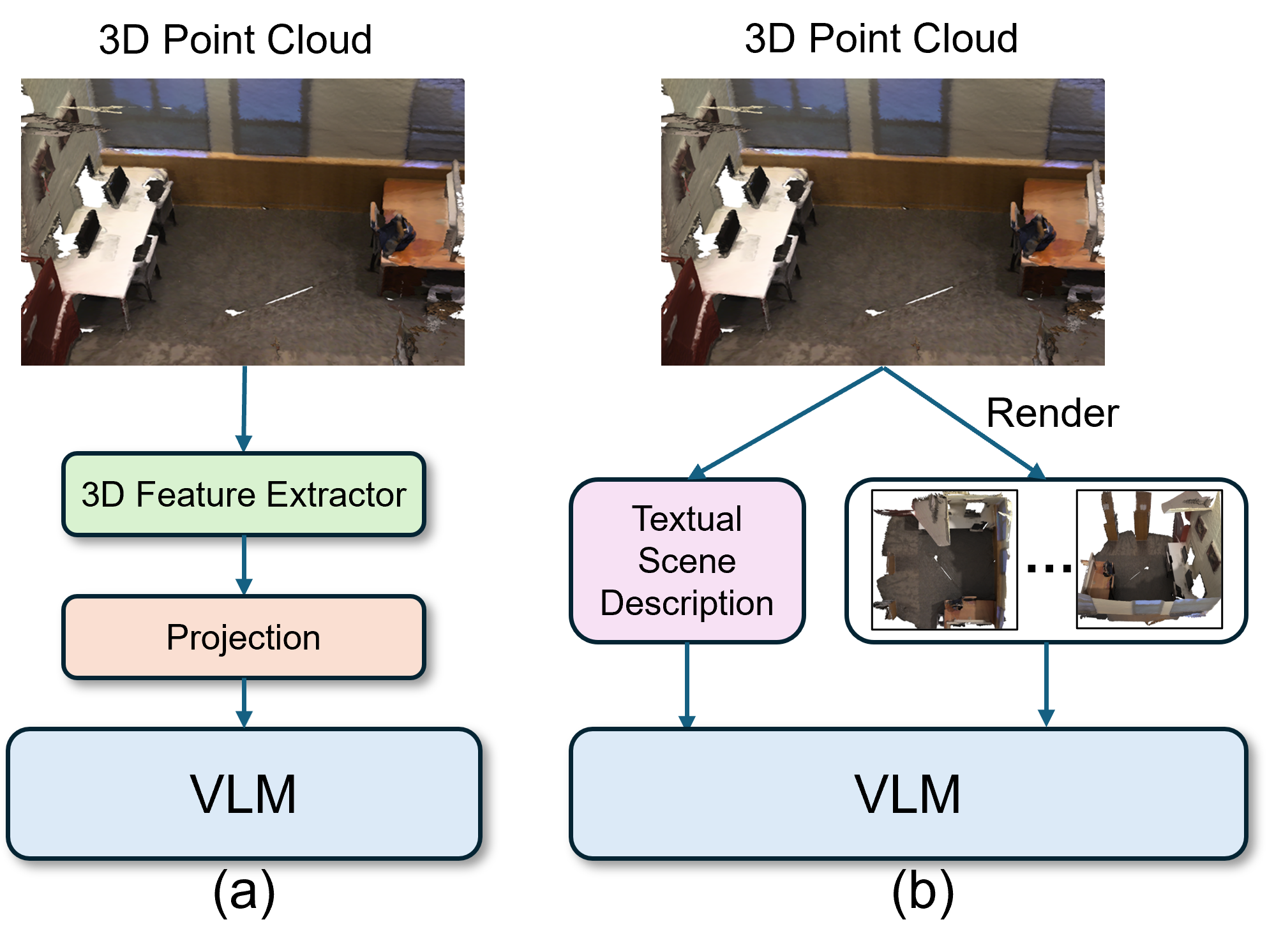}
    \caption{Comparison of (a) conventional embedding space alignment, which maps 3D point cloud features into the VLM via feature extraction and projection, and (b) our proposed input space alignment, which converts the 3D scene into textual descriptions and multi-view rendered images before feeding them directly into the VLM.}
    \label{fig:Intro}
\end{figure}

3D scene understanding has gained increasing attention due to its foundational role in embodied AI, metaverse, robotics, and beyond. While recent advances in vision-language models (VLMs) \cite{Qwen2, li2023blip, alayrac2022flamingo, zhu2023minigpt, liu2023visual} have revolutionized reasoning in 2D vision-language tasks, extending these models to 3D scene understanding remains highly challenging. Existing approaches \cite{huang2024chat, man2024situational, wang2023chat, zhang2024spartun3d, fu2024scene} typically attempt to bridge 3D and language modalities by projecting 3D features into the VLM’s embedding space via projection or adaptation layers as shown in Fig.~\ref{fig:Intro}(a), aiming to leverage the VLM’s extensive prior knowledge. However, this implicit feature alignment is limited because of the lack of large-scale annotated 3D datasets, and the intricate and high-dimensional nature of spatial relationships in 3D space, which results in suboptimal performance. 

To address these limitations, we introduce a hierarchical multimodal representation that enables explicit input-level alignment between 3D scenes and VLMs. Unlike previous approaches that map 3D features into the VLM’s latent space, our method harnesses the complementary strengths of language and 2D visual evidence. Concretely, we first transform the 3D scene into a detailed textual description that enumerates all detected objects and encodes their spatial locations in natural language. This approach offers a critical advantage: by representing 3D scene as interpretable language, we allow the VLM to reason within its native modality, leveraging its extensive pretraining on textual data. In contrast, aligning 3D features with latent language spaces typically introduces an opaque mapping, where geometric information might be lost or rendered inaccessible to the model. Our explicit linguistic representation thus provides fine-grained spatial context in a form that the VLM is inherently equipped to process, enhancing its ability to reason about complex 3D scene layouts.

Recognizing that text descriptions alone may not fully encapsulate the richness of real-world 3D scenes, we augment our representation with multi-view images that provide complementary visual context. Each scene is rendered from multiple perspectives, including a bird's-eye view and four top-down directions (forward, left, right, and backward), ensuring that all information is visually accessible to the model. One limitation of directly using the image encoder of VLMs is that the image features are encoded patch-wise, which might not capture the global information of the scene. 
To overcome this, we further introduce a hierarchical aggregation strategy: patch-wise features are aggregated into learnable view-level and scene-level tokens using attention mechanisms, allowing the model to jointly reason over both local details and global context. This hierarchical representation is critical for handling the diverse question types encountered in 3D scene Q\&A, ranging from object-specific queries to those requiring a global understanding of the scene layout.

We conduct comprehensive experiments on two challenging benchmarks: situated 3D Q\&A dataset SQA3D \cite{SQA3D}, which focuses on agent-centric, situated reasoning, and general 3D Q\&A dataset ScanQA \cite{ScanQA}, which covers a broader range of questions that require 3D scene understanding. Our approach consistently achieves state-of-the-art performance, outperforming existing methods by a substantial margin. Qualitative and quantitative analyses further demonstrate the effectiveness of our hierarchical multimodal representation. In summary, our contribution can be summarized as follows: 
\begin{itemize}
\item We introduce a multimodal representation for 3D scene understanding that explicitly aligns 3D data with VLMs at the input level, combining text descriptions with multi-view visual evidence for enhanced spatial reasoning.
\item We propose a hierarchical feature aggregation framework that captures both local and global scene context through learnable view-level and scene-level tokens.
\item Our approach sets a new state-of-the-art on both situated and general 3D Q\&A benchmarks, highlighting the effectiveness of our approach.
\end{itemize}

\section{Related Work}
\label{sec:Related Work}

\paragraph{3D Scene Understanding} 3D scene understanding has long been a fundamental challenge in computer vision, primarily due to the complex spatial relationships and high-dimensional structures inherent to 3D environments, which pose greater difficulty compared to their 2D counterparts. Early research in this area focused on core perception tasks such as 3D classification \cite{qi2017pointnet, qi2017pointnet++}, object detection \cite{qi2019deep}, and semantic segmentation \cite{thomas2019kpconv}. More recently, there has been a surge of interest in language-based 3D reasoning. Notable directions include 3D visual grounding \cite{achlioptas2020referit3d, chen2020scanrefer, huang2022multi}, which seeks to localize objects within 3D scenes based on textual queries; 3D dense captioning \cite{chen2023end, chen2024vote2cap, chen2021scan2cap}, which generates descriptive sentences for localized regions in 3D space; and 3D question answering \cite{ScanQA, parelli2023clip, luo2025dspnet, huang2025unveiling}, which involves answering free-form questions grounded in 3D data. Situated 3D question answering \cite{SQA3D, man2024situational} further extends these challenges by requiring context-aware reasoning from the perspective of embodied agents. Due to the scarcity of 3D-language data, most approaches leverage the semantic priors of VLMs for various 3D tasks. One typical way to align 3D features with VLMs is via feature fusion or projection layers. However, such implicit alignment is fundamentally limited by the inherent complexity of modeling fine-grained spatial relationships, often resulting in suboptimal reasoning performance. In this work, we address these limitations by proposing an explicit, input-level alignment strategy that combines text descriptions and multi-view visual evidence for more effective 3D scene reasoning.

\paragraph{Vision Language Models} The advent of VLMs \cite{zhu2023minigpt, Qwen2, alayrac2022flamingo, li2023blip, liu2023visual} has led to remarkable progress in integrating vision and language across a variety of domains. However, extending these advances to the 3D domain remains challenging, largely due to the scarcity of large-scale 3D scene-text datasets. To mitigate this, recent approaches \cite{man2024situational, wang2023chat, huang2024chat, zhang2024spartun3d, fu2024scene} seek to harness the prior knowledge of pretrained VLMs by adapting them to 3D data through various alignment strategies. For example, 3D-LLM \cite{hong20233d} constructs 3D scene representations from 2D pretrained features of rendered multi-view images, mapping the resulting 3D features into the same embedding space as 2D features. Chat-3D \cite{wang2023chat} adopts a three-stage training scheme, leveraging 3D scene-text data to progressively align object attributes and spatial relationships with language models. Spartun3D \cite{zhang2024spartun3d} introduces a 3D object-text alignment loss that guides point cloud encoder training using detailed textual descriptions for individual objects within a scene. Despite these advances, challenges remain in precisely grounding objects and modeling complex spatial relationships within 3D environments. In this work, we propose to leverage the prior knowledge of VLMs by explicitly aligning 3D data with LLMs at the input level, facilitating more effective 3D scene reasoning.

\begin{figure*}[tbh]
    \centering
    \includegraphics[width=\textwidth]{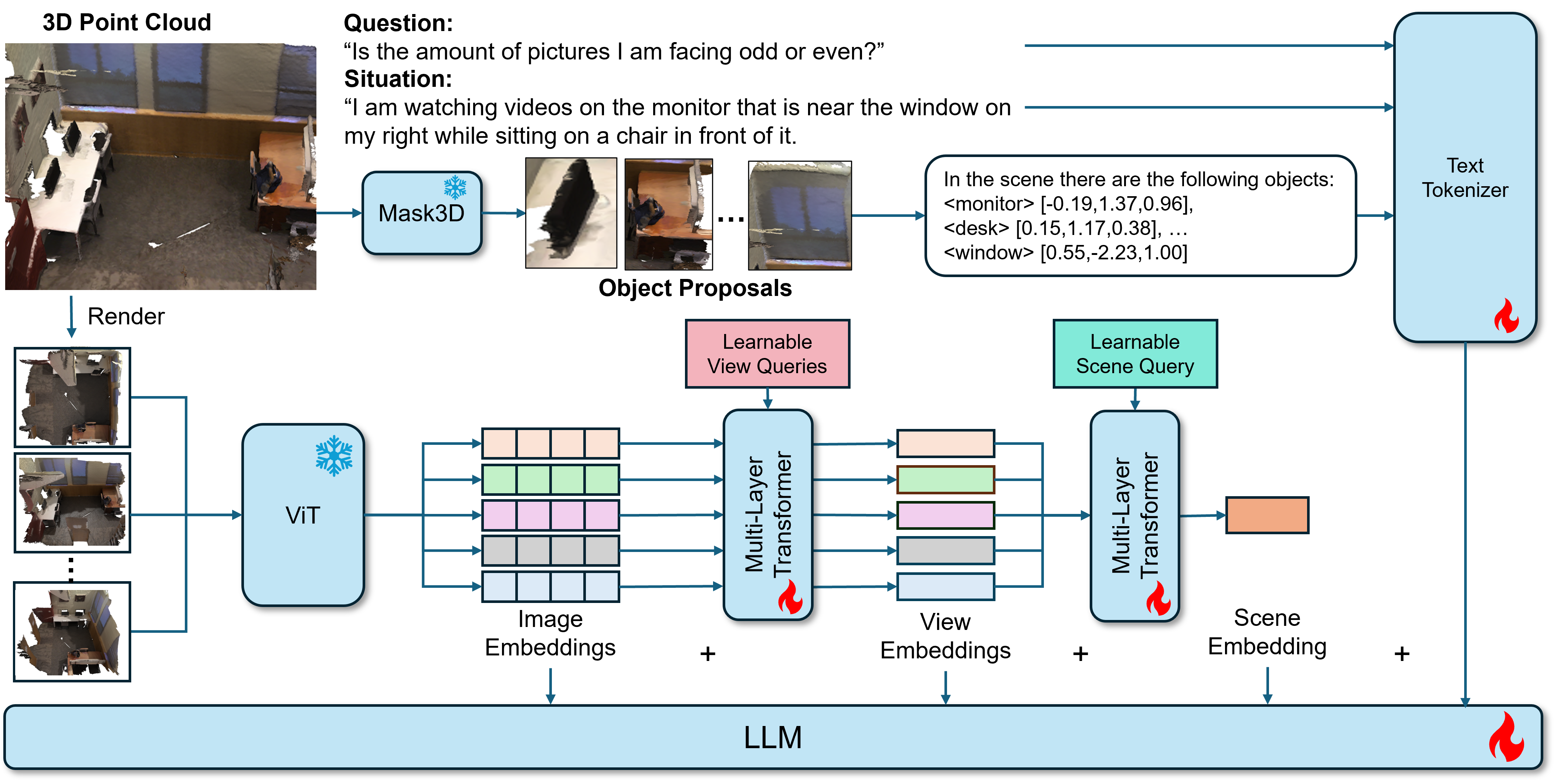}
    \caption{Overview of our proposed pipeline, which consists of two main stages: (1) generation of structured text descriptions from the 3D scene and (2) multi-view image rendering with hierarchical visual feature extraction. Both the textual and visual information are jointly fed into model to generate the final answer.}
    \label{fig:architecture2}
\end{figure*}

\section{Method}
\subsection{Overview}

We propose a hierarchical multimodal representation to explicitly align 3D scene information with VLMs, thereby leveraging the rich prior knowledge of pretrained VLMs for complex 3D scene question answering. Our representation integrates scene text descriptions with multi-view visual evidence, allowing the model to jointly reason about both spatial relationships and appearance-based cues in 3D environments. Unlike prior approaches that map 3D features into the VLM embedding space via projection layers, our method achieves alignment directly at the input level, substantially enhancing spatial grounding. Specifically,
given a 3D point cloud $\mathcal{P}$ and a natural language question $q^{\mathrm{txt}}$, our goal is to answer $q^{\mathrm{txt}}$ using all information present in the scene. For situated 3D Q\&A tasks, an additional situation description $s^{\mathrm{txt}}$ is provided, indicating the agent's location or perspective. The overall pipeline of our approach is shown in Fig.~\ref{fig:architecture2}, which consists of two key stages: (1) text description generation from 3D scene, (2) multi-view image rendering and hierarchical feature extraction. This design enables the model to flexibly attend to both local and global scene context.

\subsection{3D Scene Text Description Generation}

Existing approaches for 3D Q\&A typically attempt to align 3D point cloud features with the VLM feature space using learned projections. In contrast, we introduce a semantic scene-to-text translation step, leveraging the language-centric training of modern VLMs. Specifically, we employ an off-the-shelf 3D instance segmentation model Mask3D~\cite{mask3d} to generate a comprehensive set of object proposals from the point cloud. To reduce redundancy, we prune the set of detected objects using non-maximum suppression based on bounding box IoU, as well as majority voting in overlapping regions. The resulting object list is then used to construct a 3D scene textual description $t^{\textrm{3D}}$, which details the object classes, their spatial coordinates. The format is designed to be object-centric and spatially explicit, for example:
\begin{quote}
In the scene there are the following objects: \textless monitor\textgreater{} at [-0.19, 1.37, 0.96], \textless desk\textgreater{} at  [0.15, 1.17, 0.38] , \textless window\textgreater{} at [0.55, -2.23, 1.00].
\end{quote}
This explicit textual grounding provides the VLM with interpretable, language-native spatial information, improving the model's ability to reason about object locations, relations, and overall scene layout.

\subsection{Hierarchical Visual Representation}
While scene text descriptions capture spatial relationships, they may omit essential visual attributes such as color, texture, or fine-grained object appearance. To address this, we complement the textual modality with a hierarchical visual representation derived from multi-view images. The multi-view images are rendered from the 3D point cloud from a bird's-eye view and four top-down directions (forward, left, right, and backward), ensuring comprehensive coverage and minimizing occlusion--example is shown in Fig.~\ref{fig:multiview}. More details on the multi-view image rendering are provided in the Implementation Details.
Each image is encoded by a pretrained vision transformer $E_v$, yielding patch-level features $f_i^m = E_v(I_i^m)$, where $m$ indexes the view and $i$ the patch. To further enable coarse-to-fine reasoning, we aggregate patch features within each view using a learnable cross-attention mechanism to obtain view-level tokens:
\begin{equation}
V^m = \mathrm{Attention}(Q=q^m, K=f_i^m, V=f_i^m),
\label{eq:view-level token}
\end{equation}
where $q^m$ is a learned query embedding specific to each view. Subsequently, all view-level tokens are aggregated via another cross-attention layer, using a global scene query $q$, to produce a scene-level representation:
\begin{equation}
S = \mathrm{Attention}(Q=q, K=V^m, V=V^m).
\label{eq:scene-level token}
\end{equation}
This hierarchical aggregation enables the model to represent both localized details and holistic scene structure, which is critical for answering questions requiring either fine-grained or global understanding.

The final hierarchical visual embedding $F_v$ is constructed as:
\begin{equation}
F_v = \mathrm{Concat}(f_1^1, f_2^1, ...,f_N^5, V^1, ...,V^5, S).
\end{equation}
To enhance interpretability and provide explicit structure to the language model, we wrap the visual features with specialized demarcation tokens: \textless\textbar vision\_start\textbar\textgreater{} and \textless\textbar vision\_end\textbar\textgreater{} . Additionally, we introduce new tokens: \textless\textbar view\_start\textbar\textgreater{}, \textless\textbar view\_end\textbar\textgreater{}, \textless\textbar scene\_start\textbar\textgreater{}, and \textless\textbar  scene\_end\textbar\textgreater{}. This explicit marking allows the model to distinguish between different levels of abstraction within the visual modality. 

\subsection{Training and Optimization}
Given the generated 3D scene text description $t^{\textrm{3D}}$, hierarchical visual tokens $F_v$, the natural language question $q^{\mathrm{txt}}$, and (if available) situated context $s^{\textrm{txt}}$, we concatenate these modalities as input to the LLM. Note that all text modality are tokenized before feeding into LLM. Finally, the model is trained to generate the response sequence $\hat{A} = [\hat{a}_1, \hat{a}_2, \ldots, \hat{a}_L]$, where $L$ is the length of the predicted response. The model predicts answers in an autoregressive manner, where each token of the answer is predicted conditioned on all previous tokens and the multimodal context. The loss for generating the $i$-th answer token is defined as:
\begin{equation}
\mathcal{L}(\theta) = -\sum_{i=1}^{L} \log p_\theta(\hat{a}_i \mid \hat{a}_{<i}, t^{3D}, F_v, q^{\mathrm{txt}}),
\end{equation}
where $\theta$ denotes all trainable parameters in the model during finetuning, including (1) the weights of the cross-attention layers for hierarchical visual aggregation, (2) the learnable query tokens for both view-level and scene-level feature extraction, (3) embedding layer to learn new special tokens that demarcates the new representation and (4) selected layers in LLM.
More details about the specific fine-tuning layers are provided in the Training Details.

\section{Experiments}
\paragraph{Training Details}

We finetune the pretrained Qwen-VL 2.5 7B Instruct vision-language model \cite{Qwen2} for the 3D question answering task using the Low-Rank Adaptation (LoRA) \cite{hu2022lora} technique. Specifically, LoRA adapters are inserted into the \texttt{q\_proj}, \texttt{k\_proj}, \texttt{v\_proj}, and \texttt{o\_proj} modules of the model’s attention layers, enabling efficient adaptation with a minimal number of trainable parameters. The vision encoder backbone (ViT) is kept frozen throughout training to preserve the pretrained visual representations and reduce computational overhead. We use the AdamW optimizer with a base learning rate of $1\times10^{-4}$, batch size is set to 1 with a gradient accumulation step of 16. All experiments are conducted on a single Nvidia A100 80GB GPU. Unless otherwise specified, all other hyperparameters follow the defaults provided by the Qwen-VL 2.5 implementation.

\begin{figure*}[tbh]
    \centering
    \includegraphics[width=\textwidth]{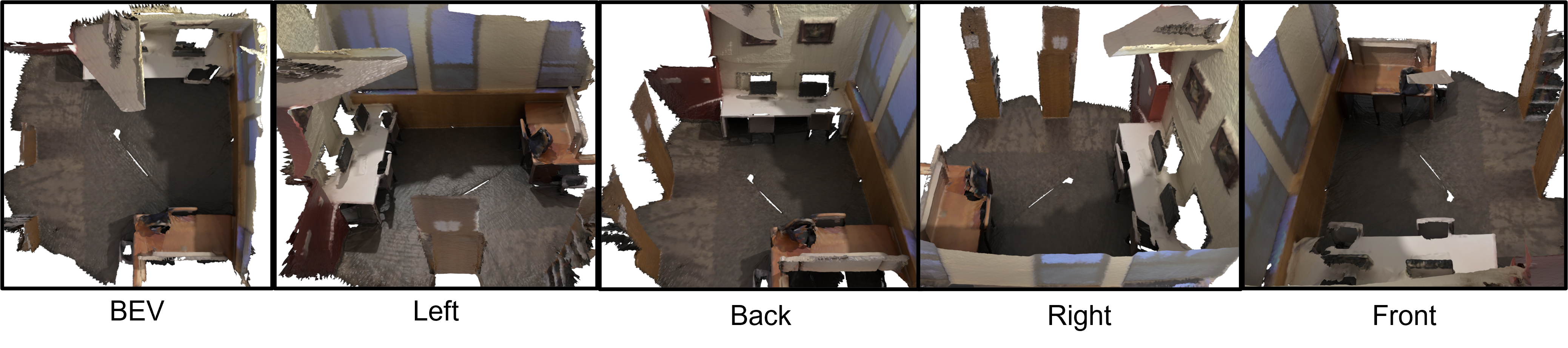}
    \caption{An example of the rendered multi-view images, including a bird's-eye view and four top-down directions (forward, left, right, and backward) to ensures comprehensive spatial coverage of the 3D scene.}
    \label{fig:multiview}
\end{figure*}

\paragraph{Implementation Details} For each scene, we generate multi-view 2D images by strategically positioning a virtual camera at multiple viewpoints within the 3D environment. To obtain a holistic perspective, we first place the camera directly above the scene and orient it downward, capturing a bird’s-eye view that highlights the overall spatial layout. To further enrich contextual understanding and mitigate occlusion, we render four additional images by positioning the camera at offset locations along the front, back, left, and right cardinal directions, each angled obliquely toward the center of the scene. This multi-view setup ensures comprehensive spatial coverage and maximizes the visibility of objects from diverse perspectives. 

An example of the rendered multi-view images are shown in Fig.~\ref{fig:multiview}.
We adopt Qwen’s dynamic resizing to preserve aspect ratio and constrain each image to about 500 visual tokens. On average, the total number of tokens, including system prompts, images, multi-level representations, and scene descriptions, is 4,110. 
To effectively integrate multi-view visual information, we modify the visual-textual interface of the VLM by 
introducing an intermediate hierarchical representation between the vision encoder and the language model. Specifically, we employ two transformer layers, each with four attention heads, to compute hierarchical visual representations at the view and scene levels. First, a set of learnable query tokens performs cross-attention over the patch embeddings generated by the ViT encoder for each image, yielding five compact view-level representations. Subsequently, a second learnable query token attends over these five view tokens, producing a unified scene-level embedding that captures global context. 
\begin{figure}[htbp]
\begin{tcolorbox}[colback=lightgray!10,
                    colframe=black,
                    width=\textwidth,
                    arc=1mm, auto outer arc,
                    boxrule=0.5pt,
                    ]
\textbf{System}: You are a assistant that can understand a scene, you will be provided with images of top-down views of the scene,view representations and scene representation, a situation and coordinates[x,y,z] of objects of the scene. Answer the question using a single word or phrase 

\textbf{User:} \texttt{<image>}\texttt{<image>}\texttt{<image>}\texttt{<image>}\texttt{<image>}\texttt{<view>}\texttt{<view>}\texttt{<view>}\texttt{<view>} \texttt{<scene>}

Situation: I am facing a window and there is a desk on my right and a chair behind me.

In the scene there are the objects:\textless plant\textgreater[-0.7083,1.6266,0.9519],\textless plant\textgreater[-0.5052,1.6281,0.9397],\textless radiator\textgreater[1.0013,1.6693,0.2507],\textless window\textgreater[0.1105,1.4761,1.6420]...

Question: What color is the desk to my right?

\end{tcolorbox}
\caption{Prompt template for our model}
\label{tab:prompt_input}
\end{figure}

\paragraph{Datasets}
We evaluate our proposed method on two widely used benchmark datasets: SQA3D~\cite{SQA3D} and ScanQA~\cite{ScanQA}.
\textbf{SQA3D} consists of over 33,000 question-answer pairs specifically curated for 3D visual question answering, alongside 26,000 unique situational descriptions designed for agent-centric situation estimation. Each data instance comprises a 3D point cloud scene, a free-form situational description, a natural language question, and answer annotations. This dataset emphasizes the ability to perform complex spatial reasoning and interpret situational context within 3D environments.
\textbf{ScanQA} contains more than 41,000 question-answer pairs grounded in diverse 3D indoor scenes. Unlike SQA3D, ScanQA primarily targets general 3DQA, challenging models to reason about scene layouts, object attributes, and semantic relationships solely from 3D data and associated questions.
For both datasets, we follow the standard training, validation, and test splits as provided in their official protocols. 

\paragraph{Evaluation Metrics}
We assess model performance on SQA3D using the Exact Match at Top-1 (EM@1) metric, following the standard evaluation protocol established in~\cite{SQA3D}. EM@1 measures the percentage of predictions that exactly match the reference answer and is equivalent to Top-1 answer accuracy. To provide further insight into the model’s capabilities, we also report EM@1 scores broken down by question type, categorizing questions according to their leading word (e.g., “What,” “Is,” “How,” “Can,” “Which,” and “Other”).
For ScanQA, we evaluate with BLEU \cite{papineni2002bleu}, ROUGE \cite{lin2004rouge}, METEOR \cite{banerjee2005meteor}, and CIDEr \cite{vedantam2015cider}. These metrics evaluate the overlap, semantic similarity, and fluency of the predicted answers relative to the ground truth.

\subsection{Results for Situated 3D Q\&A}
\label{sec:results_situated_qa}

We evaluate our proposed method on the SQA3D dataset under two experimental settings, following the protocol in~\cite{man2024situational}. \textbf{(a) Text-Only Situation:}
In the first setting, only the textual situation description is available—ground truth location and orientation information are not provided. Here, the free-form situational text $s^{\mathrm{txt}}$ is concatenated with the generated 3D scene text description $t^{3D}$, hierarchical visual tokens $F_v$, and the natural language question $q^{\mathrm{txt}}$, and then fed as input to the VLM. This setting simulates a scenario where only agent-centric language descriptions of position are accessible. \textbf{(b) Text + Ground Truth Location:}
In the second setting, both the textual situation and ground truth location and rotation are available. In this case, we further enrich the 3D scene text description by incorporating explicit spatial relationships between each object and the agent’s current location. These relationships are computed based on the bounding box coordinates of each object relative to the agent, and are expressed using a clock-wise notation. For example:
\begin{quote}
``To my 12 o'clock there is a \textless monitor\textgreater [-0.19, 1.37, 0.96]. To my 2 o'clock there is a \textless desk\textgreater [0.15, 1.17, 0.38], and \textless window\textgreater [0.55, -2.23, 1.00] ...''
\end{quote}
This explicit geometric grounding helps the model reason about object positions and the global structure of the scene with greater precision. We denote this setting with asterisk mark (*) in our results.

The results for SQA3D are presented in Tab.~\ref{tab:sqa3d_em}, where baseline results directly reported from SIG3D except for Chat-Scene and Spartun3D-LLM. Our proposed approach achieves a substantial performance improvement over existing methods. Notably, our method outperforms the text-3D fusion based SIG3D, the projection-layer based approach Chat-Scene, and the explicit 3D-text alignment based method Spartun3D-LLM. It is worth mentioning that Spartun3D-LLM is built upon the LEO~\cite{huang2023embodied} backbone, which is extensively pretrained on a wide range of 3D tasks. These results strongly demonstrate the effectiveness of explicit input-level alignment, as proposed in our method, for enabling VLMs to better reason over complex 3D scenes and spatial relationships.

\begin{table*}
\small
\setlength{\tabcolsep}{4pt}
\centering
\begin{tabular}{lc@{\hspace{8mm}}c@{\hspace{6mm}}c@{\hspace{6mm}}c@{\hspace{6mm}}c@{\hspace{6mm}}c@{\hspace{6mm}}c}
\toprule
\multirow{2}{*}{\textbf{Model}} & \multicolumn{6}{c}{\textbf{Question Breakdown}} & \multirow{2}{*}{\textbf{Overall}}\\
\cmidrule(r{2mm}){2-7}
\
&\textbf{What} & \textbf{Is} & \textbf{How} & \textbf{Can} & \textbf{Which} & \textbf{Other} \\
\midrule
GPT-3~\cite{brown2020language}                          & 39.7 & 46.0 & 40.5 & 45.6 & 36.1 & 38.4 & 41.0\\
ClipBERT~\cite{lei2021clipbert}                         & 30.2 & 60.1 & 38.7 & 63.3 & 42.5 & 42.7 & 43.3\\
MCAN~\cite{yu2019mcan}                                  & 28.9 & 59.7 & 44.1 & 68.3 & 40.7 & 40.5 & 43.4\\
ScanQA~\cite{ScanQA}                                    & 28.6 & 65.0 & 47.3 & 66.3 & 43.9 & 42.9 & 45.3\\
SQA3D~\cite{SQA3D}                                      & 33.5 & 66.1 & 42.4 & 69.5 & 43.0 & 46.4 & 47.2\\
Multi-CLIP~\cite{delitzas2023multiclip}                 & -    & -    & -    & -    & -    & -    & 48.0\\
LM4Vision~\cite{pang2023lm4vision}                      & 34.3 & 67.1 & 48.2 & 68.3 & 48.9 & 45.6 & 48.1\\
3D-LLM~\cite{hong20233d}                                & 36.5 & 65.6 & 47.2 & 68.8 & 48.0 & 46.3 & 48.1\\
3D-VisTA~\cite{zhu20233dvista}                          & 34.8 & 63.3 & 45.4 & 69.8 & 47.2 & 48.1 & 48.5\\
SIG3D \cite{man2024situational}                         & 35.6 & 67.2 & 48.5 & 71.4 & 49.1 & 45.8 & 52.6\\
SIG3D * \cite{man2024situational}                       &-  &- &- &- &- &-                        & 53.9\\
Chat-Scene~\cite{huang2024chat}                         &-  &- &- &- &- &-                        & 54.6\\
Spartun3D-LLM~\cite{zhang2024spartun3d}                 &-  &- &- &- &- &-                        & 54.8\\
DSPNet~\cite{luo2025dspnet}                             & 38.2 & 66.0 & 51.2 & 66.6 & 42.5 & 51.6 & 50.4 \\
\midrule
Ours                                                    & 47.2 & 71.3 & 57.0 & 72.5 & 63.5 & 56.0 & 58.4 \\    
Ours *                                                  & 51.9 & 73.2 & 56.8 & 74.0 & 69.1 & 63.1 & 62.1 \\
\bottomrule
\end{tabular}
\caption{Quantitative results on the SQA3D benchmark. * denotes the setting where ground truth situation is available. Results are reported on EM@1 metric.}
\label{tab:sqa3d_em}
\end{table*}

\subsection{Results for General 3D Q\&A}

We report the results of our approach and baseline methods on the ScanQA dataset in Tab.~\ref{tab:scanqa}. Results for all methods are taken from SIG3D, except for LEO, Scene-LLM, and Chat-Scene, which are reported from the Chat-Scene paper. Our approach achieves significant improvements over existing methods across most evaluation metrics. We note, however, that the BLEU-4 score is marginally lower than that of Chat-Scene. One possible explanation is that Chat-Scene may generate answers with longer exact n-gram overlaps to the ground truth, leading to higher BLEU-4. Nevertheless, our method surpasses all baselines for the remaining metrics, demonstrating the effectiveness of our approach to general 3D Q\&A task.

\begin{table}
\setlength{\tabcolsep}{.7mm} 
\centering
\small
\begin{tabular}{lccccc}
\toprule
\textbf{Model}                                  & \textbf{B-1}  & \textbf{B-4}   & \textbf{R}   & \textbf{M}    & \textbf{C}    \\
\midrule
BLIP2~\cite{li2023blip}                         & 29.7 & 5.9  & 26.6 & 11.3 & 45.7  \\
Flamingo~\cite{alayrac2022flamingo}             & 25.6 & 8.4  & 31.1 & 11.3 & 55.0  \\
VN+MCAN~\cite{yu2019mcan}                       & 28.0 & 6.2  & 29.8 & 11.4 & 54.7  \\
SR+MCAN~\cite{yu2019mcan}                       & 26.9 & 7.9  & 30.0 & 11.5 & 55.4  \\
ScanQA~\cite{ScanQA}                            & 30.2 & 10.1 & 33.3 & 13.1 & 64.9  \\
3D-LLM~\cite{hong20233d}                        & 39.3 & 12.0 & 35.7 & 14.5 & 69.4  \\
SIG3D~\cite{man2024situational}                 & 39.5 & 12.4 & 35.9 & 13.4 & 68.8  \\
LEO~\cite{huang2023embodied}                    & -    & 11.5 & -    & -    & 80.0  \\
Scene-LLM~\cite{fu2024scene}                    & -    & 12.0 & -    & -    & 80.0  \\
Chat-Scene~\cite{huang2024chat}                 & -    & 14.3 & -    & -    & 87.7  \\
\midrule
Ours                                            & 44.3 & 13.5 & 43.9 & 17.8 & 89.0  \\
\bottomrule
\end{tabular}
\caption{Quantitative results on ScanQA Dataset. We report results on metrics, BLEU-1, BLEU-4, ROUGE, METEOR and CIDEr.}
\label{tab:scanqa} 
\end{table}

\subsection{Ablation Study}
We conduct ablation study on the SQA3D dataset to evaluate the contribution of each component in our framework. In all experiments, the model receives the question and situation description as inputs; we then vary the remaining modalities as follows: 
\textbf{(a)} Multi-view Images (MV): The model has access only to multi-view images of the scene.
\textbf{(b)} Coordinate Text Description (CT): The model is provided with a textual description specifying object coordinates.
\textbf{(c)} Coordinate \& Direction Text Description (CDT): The model receives text descriptions including both object coordinates and object-relative directional cues. The object-relative directional information is represented using a clock-wise notation, as described in Sec. \ref{sec:results_situated_qa}.
\textbf{(d)} Coordinate \& Direction Text Description + Multi-view Images (CDT+MV): The model combines both textual and visual scene representations.
\textbf{(e)} Coordinate \& Direction Text Description + Multi-view Images + Hierarchical Representation (CDT+MV+HR): The model incorporates our hierarchical feature aggregation on top of both modalities.
\textbf{(f)} Zero-Shot Coordinate \& Direction Text Description + Multi-view Images (ZS-CDT+MV): The base VLM is evaluated in a zero-shot setting, isolating the effect of our proposed training and representation.

The results are summarized in Tab.~\ref{tab:Ablation}. We observe that using only multi-view images (MV) already achieves strong performance. Transforming 3D spatial information into textual form—whether with coordinates alone (CT) or enriched with directional cues(CDT)—also yields competitive results, demonstrating the model’s ability to reason over language representations. Combining textual and visual modalities (CDT+MV) produces a further boost, highlighting their complementarity: text provides explicit spatial grounding, while images supply rich appearance information that is difficult to capture in language alone. Incorporating the hierarchical visual representation (CDT+MV+HR) delivers the best overall performance, confirming the importance of multi-level feature aggregation. Finally, the zero-shot baseline (ZS-CDT+MV) yields substantially lower results, confirming that our training and alignment strategy—not just the backbone model—is responsible for the performance gains.

Like all object-centric representations, our method relies on object locations derived from 3D segmentation results. To evaluate the sensitivity of our approach to segmentation quality, we conduct an ablation study comparing model performance using ground truth segmentations versus predicted segmentations generated by Mask3D. As reported in Tab.~\ref{tab:Segmentation}, we observe a performance drop when using Mask3D predictions compared to ground truth. This highlights the importance of accurate 3D segmentation for downstream reasoning. Improving robustness to segmentation errors remains an important direction for future work.

\begin{table}[t]
\small
\setlength{\tabcolsep}{2mm}  
\centering
\begin{tabular}{l@{\hspace{40mm}}c}
\toprule
\textbf{Ablation}                                                & \textbf{Overall} \\
\midrule
MV                                                               & 55.5 \\
CT                                                               & 57.2 \\
CDT                                                              & 59.8 \\
CDT + MV                                                         & 61.9 \\
CDT + MV + HR                                                    & 62.1 \\
ZS-CDT + MV                                                       & 35.5 \\
\bottomrule
\end{tabular}
\caption{Ablation study for different components.}
\label{tab:Ablation}
\end{table}
 \begin{figure*}[h!]
    \centering
    \includegraphics[width=\linewidth]{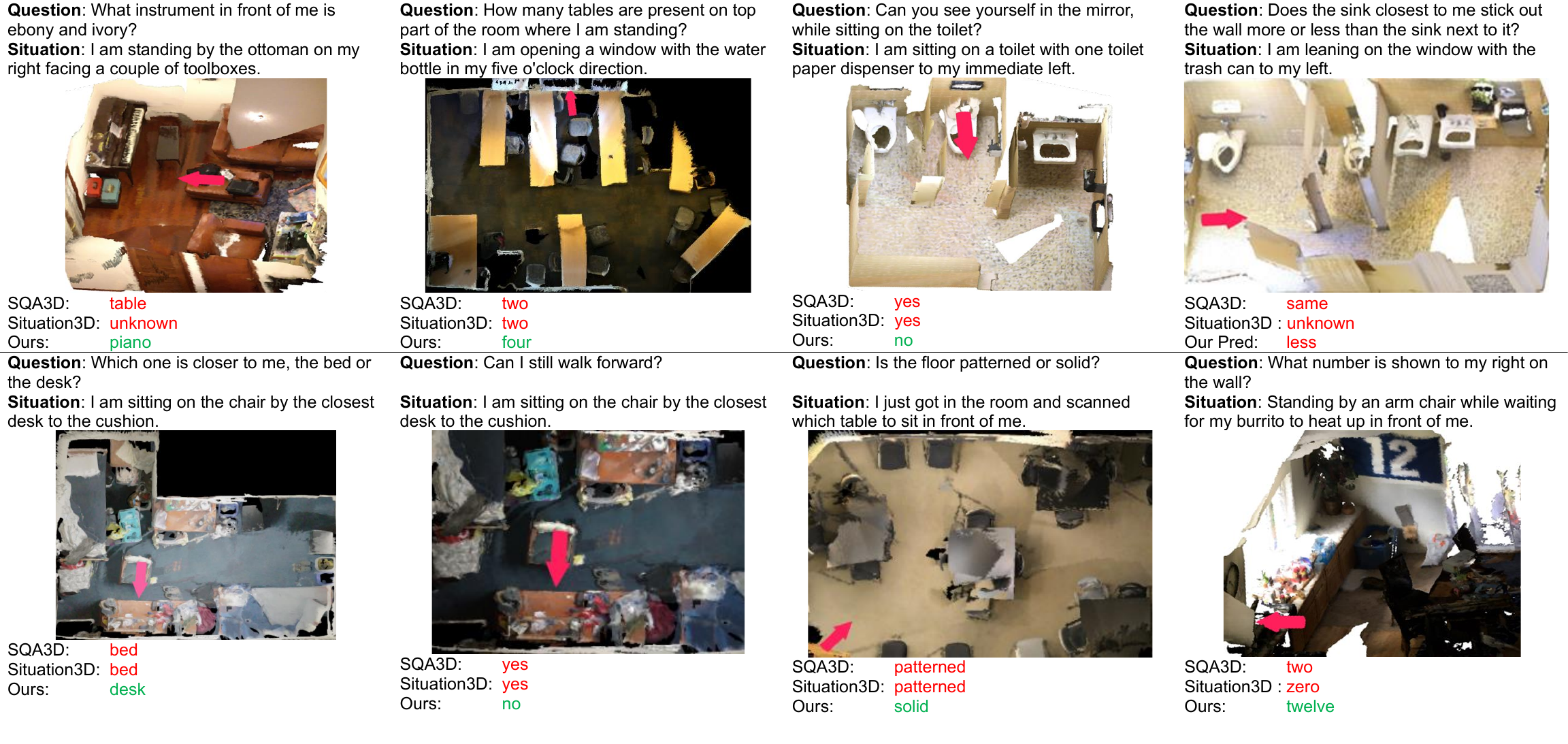}
    \caption{Qualitative examples from SQA3D dataset. The red arrow indicates the agent's position and facing direction as referenced in the situational description.}
    \label{fig:qualitative}
\end{figure*}

\begin{table}[t]
\small
\setlength{\tabcolsep}{2mm}  
\centering
\begin{tabular}{l@{\hspace{40mm}}c}
\toprule
\textbf{Segmentation}                                                & \textbf{Overall} \\
\midrule
With groundtruth                                                               & 64.2 \\
With Mask3D predictions                                                        & 58.4 \\
\bottomrule
\end{tabular}
\caption{Impact of 3D segmentation on model performance.}
\label{tab:Segmentation}
\end{table}

\subsection{Qualitative Results}

\begin{figure*}[h!]
    \centering
    \includegraphics[width=\linewidth]{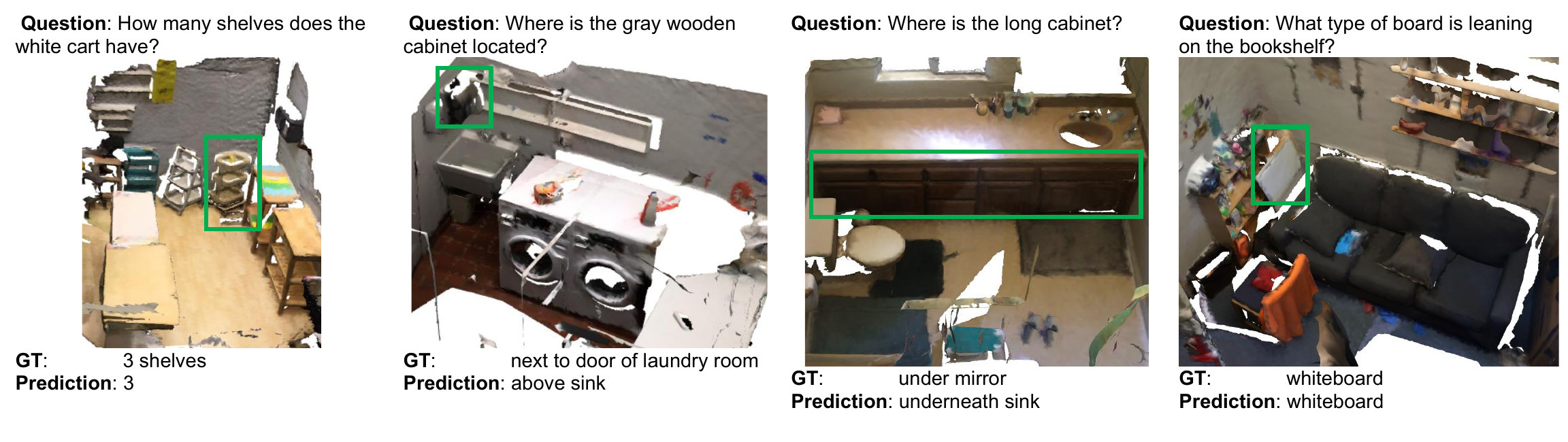}
    \caption{Qualitative examples of our method on ScanQA dataset. For illustration purposes, we include green bounding boxes that indicate the objects of interest referenced in each question.}
    \label{fig:scanqa_qualitative}
\end{figure*}

We present qualitative results on the SQA3D dataset in Fig.~\ref{fig:qualitative} to illustrate the strengths and limitations of our approach across a diverse set of question types. Compared to prior methods, our model consistently succeeds in challenging cases where others fail, demonstrating robust multimodal reasoning grounded in both textual scene descriptions and visual cues from multi-view images. By leveraging the pretrained knowledge in the language model, our approach effectively integrates object labels from the textual description with fine-grained visual features, resulting in improved object recognition and scene interpretation.
For instance, the example in the second row (third from left), the model accurately identifies subtle scene details, such as floor patterns, by combining image evidence with situational context. Additionally, the explicit encoding of object coordinates enables the model to reliably infer spatial relationships, as shown in the bottom-row example (first from left), where it correctly estimates relative object distances. Even in scenarios where direct agent location information is absent, our model is able to infer spatial context from situational and scene-level textual descriptions.

Despite strong overall performance, we acknowledge certain limitations. In particular, the model may struggle with highly complex situated questions that require multi-step spatial reasoning or nuanced understanding of agent perspective, as illustrated in the top-row example (first from the right). Addressing such challenges represents an important direction for future work.

We present qualitative examples from the ScanQA dataset in Fig.~\ref{fig:scanqa_qualitative}, where objects of interest referenced in each question are highlighted with green bounding boxes for clarity. Although the open-ended nature of the dataset means that our predictions do not always exactly match the ground truth answers, the model consistently demonstrates a strong understanding of object properties and spatial relationships. For instance, even when the predicted answer differs in phrasing (e.g., ``underneath sink" vs. ``under mirror"), the prediction remains reasonable and contextually relevant. 

\section{Conclusion} 
In this work, we introduced a hierarchical multi-modal representation that explicitly aligns 3D scene information with large vision-language models at the input level. By converting 3D scenes into text descriptions and multi-view images, and introducing a hierarchical visual aggregation mechanism, our approach enables more effective and interpretable reasoning for 3D scene reasoning. Extensive experiments on both SQA3D and ScanQA benchmarks demonstrate that our method consistently outperforms existing approaches significantly. Our results highlight the effectiveness of explicit input-level alignment for complex 3D reasoning tasks.

\paragraph{Discussion} While our approach demonstrates strong performance across multiple 3D scene understanding benchmarks, several limitations remain. Like other object-centric representations, our method depends on the quality of 3D instance segmentation or object detection; errors in these steps can propagate and impact downstream reasoning. Additionally, although input-level alignment improves spatial reasoning, the model may still struggle with questions that require subtle agent-centric understanding, particularly when situational context is ambiguous or incomplete. Addressing these challenges represents an important direction for future work.

\noindent
\paragraph{Acknowledgments} This research/project is supported by the National Research Foundation, Singapore, under its NRF Fellowship (Award\# NRF-NRFF14-2022-0001). 

\bibliographystyle{plain}
\bibliography{references}
\end{document}